\documentclass[letterpaper, 10 pt, conference]{ieeeconf}  % Comment this line out if you need a4paper

\IEEEoverridecommandlockouts                              % This command is only needed if 
\usepackage{graphicx} % for pdf, bitmapped graphics files
\usepackage{mathptmx} % assumes new font selection scheme installed
\usepackage{amsmath} % assumes amsmath package installed
\usepackage{amssymb}  % assumes amsmath package installed
\usepackage{algorithm}
\usepackage{algpseudocode}
\usepackage{cite}
\usepackage{tikz}
\usepackage{refstyle}
\usepackage{multicol}
\usepackage{amssymb}
\usepackage{array}
\usepackage{color,soul}
\usepackage{breqn}
\usepackage{multirow}
\usepackage[mathscr]{euscript}

\title{\LARGE \bf
Defo-Net: Learning Body Deformation using Generative Adversarial Networks
}

\author{Zhihua Wang*$^{1}$, Stefano Rosa*$^{1}$, Linhai Xie$^{1}$, Bo Yang$^{1}$, Sen Wang$^{2}$, \\ Niki Trigoni$^{1}$ and Andrew Markham$^{1}$ % <-this % stops a space
\thanks{*These two authors contributed equally}
\thanks{$^{1}$Wang, Rosa, Xie, Yang, Trigoni and Markham are with Department of Computer Science,
        University of Oxford, Oxford OX1 3QD, United Kingdom
        {\tt\small \{firstname.lastname\} @cs.ox.ac.uk}}%
\thanks{$^{2}$Wang is with School of Engineering and Physical Sciences, Heriot-Watt University, Edinburgh EH14 4AS, United Kingdom
        {\tt\small s.wang@hw.ac.uk}}%
}

\begin{document}

\maketitle
\thispagestyle{empty}
\pagestyle{empty}

%%%%%%%%%%%%%%%%%%%%%%%%%%%%%%%%%%%%%%%%%%%%%%%%%%%%%%%%%%%%%%%%%%%%%%%%%%%%%%%%
\begin{abstract}
Modelling the physical properties of everyday objects is a fundamental prerequisite for autonomous robots.
We present a novel generative adversarial network (\textsc{Defo-Net}), able to predict body deformations under external forces from a single RGB-D image.
The network is based on an invertible conditional Generative Adversarial Network (IcGAN) and is trained on a collection of different objects of interest generated by a physical finite element model simulator. \textsc{Defo-Net} inherits the generalisation properties of GANs. This means that the network is able to reconstruct the whole 3-D appearance of the object given a single depth view of the object and to generalise to unseen object configurations.
Contrary to traditional finite element methods, our approach is fast enough to be used in real-time applications.
We apply the network to the problem of safe and fast navigation of mobile robots carrying payloads over different obstacles and floor materials.
Experimental results in real scenarios show how a robot equipped with an RGB-D camera can use the network to predict terrain deformations under different payload configurations and use this to avoid unsafe areas.
\end{abstract}

%%%%%%%%%%%%%%%%%%%%%%%%%%%%%%%%%%%%%%%%%%%%%%%%%%%%%%%%%%%%%%%%%%%%%%%%%%%%%%%%
\section{Introduction}
A key requirement for autonomous mobile
robots is the ability to perceive their surroundings and model the environment in order to tackle high-level tasks, such as compliant manipulation and safe navigation. 
%deformable objects
For instance, many everyday objects that a robot would need to interact with are deformable or non-rigid.
% industrial AGVs and outdoor robots carrying payloads -> problem of estimating terrain deformation
In particular, traditional path planning approaches make the assumption that
the environment contains only rigid components and obstacles. In reality, not all objects or paths are rigid. Without an understanding of the potential deformation of the terrain, a wheeled robot could get stuck in a soft material or unsafely overload a weak object. To tackle this problem, the robot needs to be able to predict the traversability of the terrain, a process called \emph{terrain assessment}.
The problem of estimating the deformation of traversable spaces is 
particularly important for mobile robots carrying payloads in partially or totally unconstrained environments, for search and rescue applications, outdoor or planetary robotics in general, and industrial applications such as Automated Guided Vehicles (AGVs).

\begin{figure}[t]
\centering
\includegraphics[width=\columnwidth]{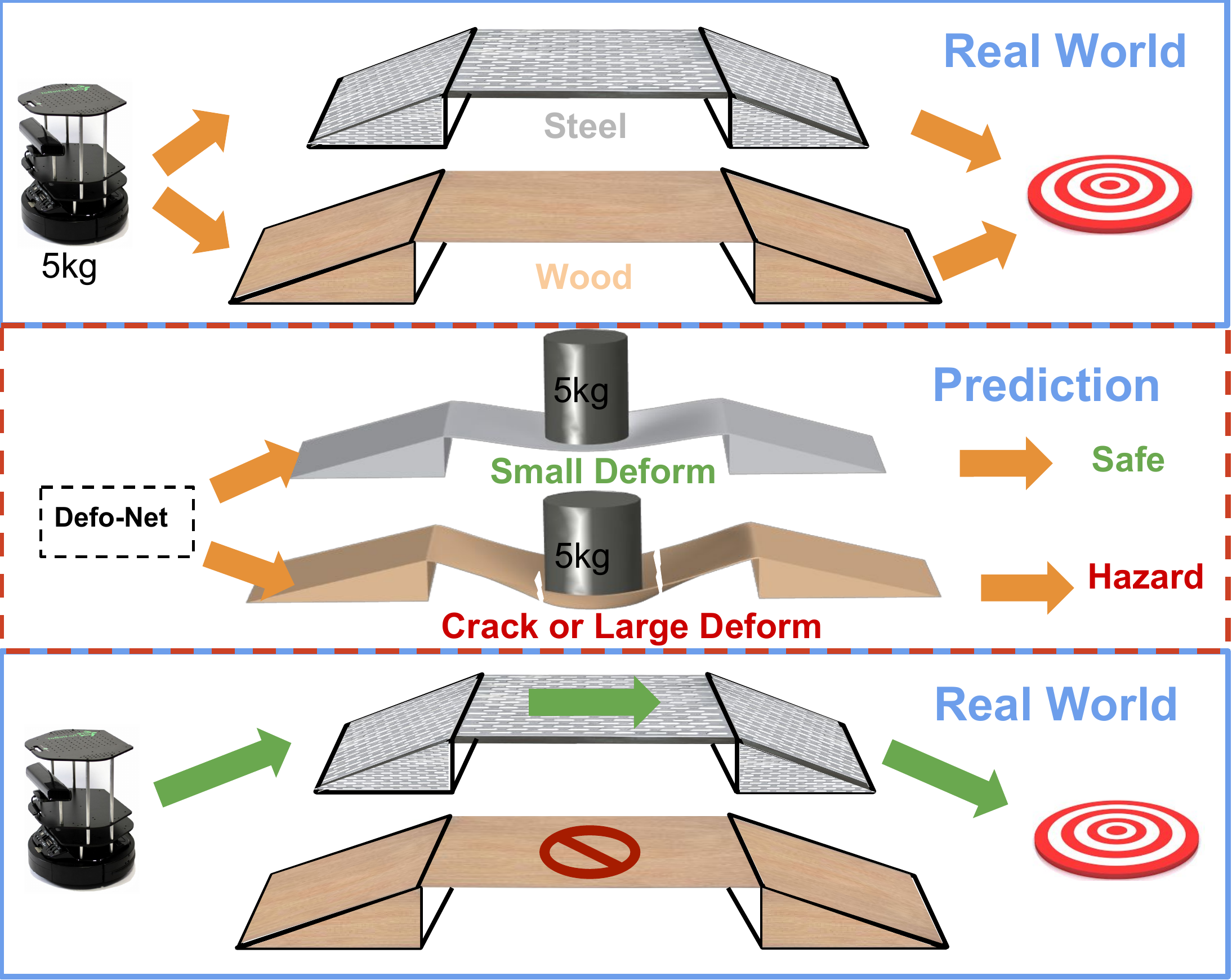}
\caption{An invertible conditional GAN is trained for predicting body deformations given an external force acting on it. This is used to estimate the 3-D deformation of potential routes. Top: the robot scans two possible routes it can take to reach the target. Middle: Using the inferred material, the RGB-D depth image, and the load, a prediction is made of the expected deformation. Bottom: Based on the results of the prediction, the path planner will take the steel bridge to reach the target.}
\label{fig:coverpage}
\end{figure}

Explicit modelling of material deformation usually requires extensive configuration and computational effort. In particular key material properties such as elasticity (or its inverse, rigidity) and the Young's Modulus need to be specified.
Deformations of non-rigid objects have been modelled in the past using mass-spring systems. While mass-spring systems can model large deformations with relatively little computational effort, they are non-intuitive and do not accurately model material properties. They are also difficult to expand to 3-D modelling.
Finite Element Method techniques (FEMs) on the other hand are highly accurate, but consequently computationally expensive, due to the large number of mesh nodes required to accurately model deformations. Co-rotational finite element approaches are a faster approximation of FEMs.
Haptic sensors have been used to learn physical properties of materials, such as elasticity \cite{gemici2014}, \cite{lee2015}.
Other methods for estimating properties of materials include a combination of induced vibrations and computer vision \cite{davis2015visual}.
More recently, the use of low-cost vision sensors and deep convolutional networks has been investigated.
In \cite{yumer2016learning} a convolutional autoencoder learns to deform a voxelized representation of input objects given an ``intention" such as ``make it sportier".
In \cite{1704.07854} the authors recently proposed a method for pre-computation of the dymanics of fluid spaces using implicit surfaces. 
%This work shares some similarities with out approach, since a generative network is trained to generate 2D surface deformations of a liquid  
% detection of materials
Material recognition is another problem that has been investigated using Convolutional Neural Networks \cite{bell2015material} \cite{dain17}.

% In order to cope with deformable objects in the planning
% process, such objects need to be handled by some sort of physical simulator.
% Achieving realistic simulations of object deformations is still an active area of
% research and real-time simulations are still impossible in practice.

% contribute
In this paper, we address the problem of estimating deformation of non-rigid structures, an open and under-explored topic in robotics.% from a single RGB-D camera image and a condition composed by a force acting on the body (area of application and strength) combined with the material.

We propose a novel deep network that combines an autoencoder and a conditional GAN, tightly coupled with an FEM-based physics simulator. Given a single RGB-D depth image of the deformable object (e.g. from a side scan of the object to be traversed) and conditioning input which includes the type of material (e.g. aluminium), the size of the force (e.g. 50~N), and the location of the force (e.g. 10~cm), the network is able to output a predicted 3-D deformation of the solid. This prediction can then be used by a path-planner to determine which is the fastest or safest path to take, given terrain information and robot payload. 

%The network is inspired by our recent work on 3-D reconstruction from a single depth image~\cite{yang17}. 
 % rewrite sentence

We evaluate our approach in three real case scenarios involving a mobile robot travelling over different bridge-like obstacles and soft ground, under different payload conditions.
We also show the generalisation capabilities of the network to unseen objects configurations.
Real world results agree closely with GAN predictions, showing its predictive power for deformable objects such as bridge-like structures and soft ground. 
A single network prediction is order of magnitudes faster that an equivalent FEM simulation, at the cost of lower resolution; this makes the approach useful for online evaluation during navigation.

In particular, the contributions of this paper are as follows:
\begin{itemize}
\item To the best of our knowledge, this is the first application of an invertible conditional GAN to the problem of learning body deformations in 3-D.
\item \textsc{Defo-Net} provides a fast and accurate approximation of FEM, which makes it suitable for predictive terrain assessment for autonomous robots.
\item We demonstrate through real-world experiments that we can generalize to different materials and structural configurations.
\end{itemize}

The remainder of this paper is as follows: in Section \ref{sec:relatedwork} we discuss some related work; in Section \ref{sec:network} we describe the proposed \textsc{Defo-Net}, the physics engine used for training and the training procedure;
in Section \ref{sec:experiments} we validate the approach in three experimental tests; finally, in Section
\ref{sec:conclusions} we draw conclusions and discuss future work.

%%%%%%%%%%%%%%%%%%%%%%%%%%%%%%%%%%%%%%%%%%%%%%%%%%%%%%%%%%%%%%%%%%%%%%%%%%%%%%%%
\section{Related work}
\label{sec:relatedwork}

\subsection{Deformable materials}
Traditional estimation techniques have been applied to the estimation of material deformations.
In \cite{frank14} the authors addressed the problem of autonomous navigation in presence of deformable obstacles, such as curtains, and manipulation of soft objects. The deformation model is learnt by the robot via physical interaction, first by running FEM simulations using the learnt deformations, then by approximating deformation cost functions for specific objects using Gaussian process regression.
In \cite{frank14icra} RGB-D images are used to learn the elasticity parameters of soft objects.
In \cite{navarro2013uncalibrated}, Lyapunov theory is used for estimating the deformation Jacobian matrix of compliant objects under elastic deformations.
Predicting shape deformations using computer vision has also been investigated in surgery applications, using an Unscented Kalman Filter for modelling the deformation of flexible needles inside soft tissues \cite{chevrie16}; RGB images of the needle are used in the filter updates. 
In \cite{dansereau16} periodic stimuli are applied to grasped objects using a gripper and the deformations are learned from RGB by magnifying the optical flow.
RGB-D images have been used in the past for estimating deformations using a variant of expectation maximization \cite{schulman2013tracking}.
\begin{figure*}[ht]
\centering
\includegraphics[width=\textwidth]{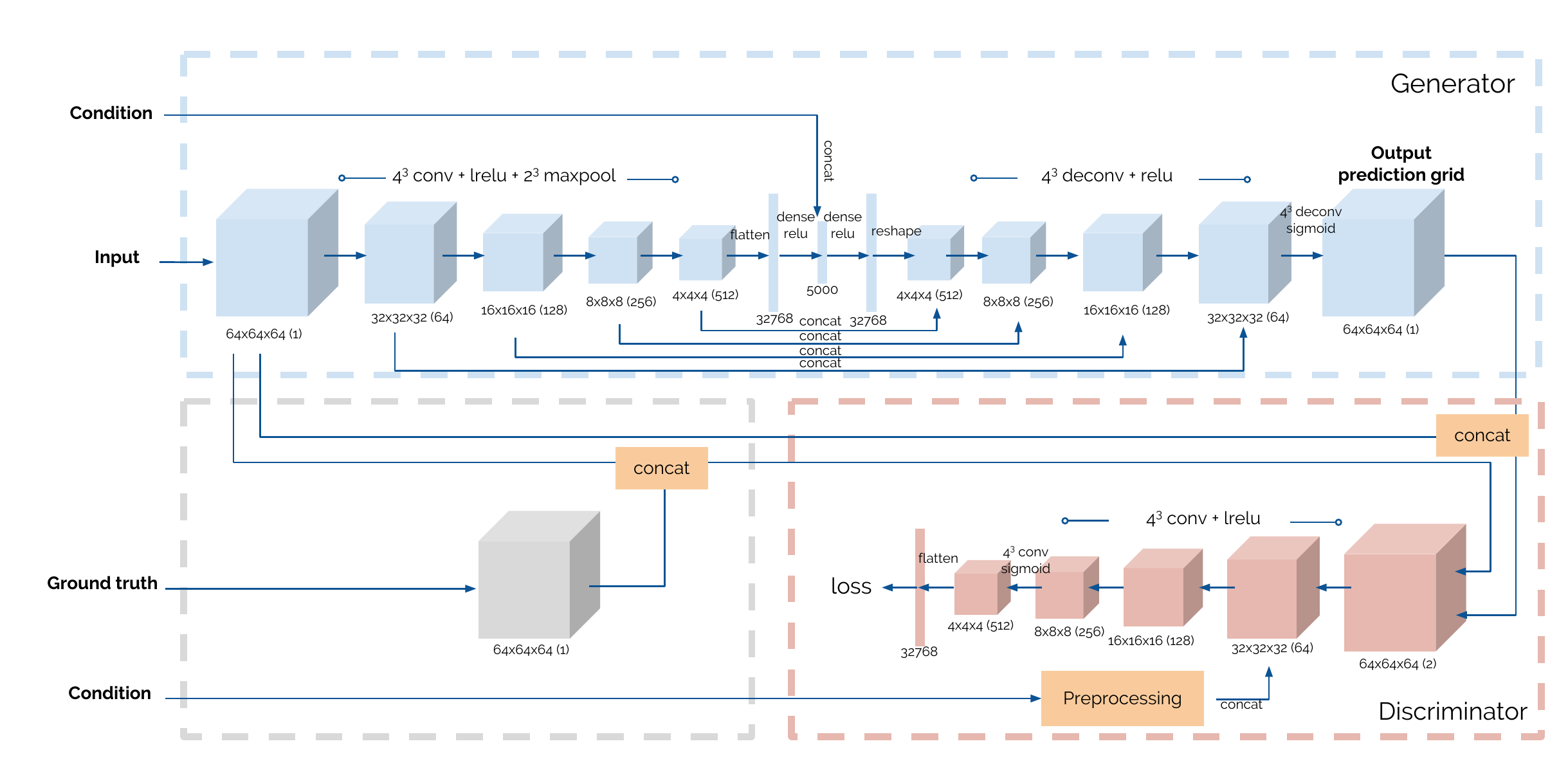}
\caption{The \textsc{Defo-Net} architecture.}
\label{fig:network}
\end{figure*}

\subsection{Intuitive physics}
%Interaction with objects
In a seminal work \cite{wu2015galileo} the authors proposed a generative model for learning physical scene understanding from video images, such as the effect of gravity and friction on objects. This is done by inverting a physics engine to obtain physical properties from observations. 
%Stability analysis
Recently, deep networks have been shown to be able to learn basic intuitive physics, such as predicting the stability of tower blocks \cite{lerer2016learning} \cite{li2016visual}, object dynamics \cite{mottaghi2016newtonian}, interacting with humans \cite{zheng2014detecting} and with other objects \cite{wu2016physics}, %Predicting the future
predicting the long-term effect of external forces \cite{mottaghi2016newtonian}, and correlating actions with effects \cite{wang2016actions}.

Predicting how actions affect the world is an open challenge. In \cite{finn2016unsupervised}
a deep model was trained in an unsupervised way to predict action-conditioned future video images of moving objects, using Convolutional Dynamic Neural Advection (CDNA) and action-conditioned LSTMs. Generative networks have been able to predict future video snippets given conditions \cite{Vondrick2016GeneratingVW}.
%fox
Recently, \textsc{SE3-Nets} \cite{byravan2017se3} were proposed %for predicting rigid body motions from 3-D point clouds. 
for learning to segment a scene into rigid objects and
predict the motion of these segmented objects under the effect of applied
forces.
The network takes as inputs a point cloud and a force vector applied to it; the network is able to segment rigid objects in the image and predict the effect of the applied force on object motion, using a layer that encodes per-pixel roto-translations.

\subsection{Generative Adversarial Networks}
Deep generative models such as Generative Adversarial Networks (GANs) \cite{goodfellow2014generative} and Variational Autoencoders (VAEs) \cite{kingma2013auto} have recently shown outstanding results in modelling high-dimensional representations and generalization abilities.
GANs have been successfully applied in different domains such as generation of text \cite{1703.00955}, learning of latent spaces \cite{chen2016infogan}, 3-D reconstruction \cite{yang17}.
% find state of the art on GANs

In the original GAN formulation the discriminator network is trained to classify real and fake examples. However, the loss function can be difficult to converge and training is often unstable.
Recently, WGAN \cite{Arjovsky2017WassersteinG} made some progress towards stable learning of GANs by using Wasserstein distance with weight clipping.
A recent work \cite{gulrajani17} proposed to penalize the norm of gradient of the discriminator with respect to its input, improving training stability.
Conditional GANs (cGANs) use external conditional information to determine specific representations of the output. Invertible cGANs (IcGANs) \cite{perarnau2016invertible}
combine an autoencoder with a cGAN and have been proved to be able to learn a good latent representation of the inputs.
cGANs have been recently used to learn mappings between input and output images
with both paired \cite{pix2pix2016} and unpaired \cite{zhu2017unpaired} images.

%\cite{radford2015unsupervised}

%%%%%%%%%%%%%%%%%%%%%%%%%%%%%%%%%%%%%%%%%%%%%%%%%%%%%%%%%%%%%%%%%%%%%%%%%%%%%%%%
\section{Network}
\label{sec:network}
Figure \ref{fig:network} shows the architecture of the proposed \textsc{Defo-Net}. It is composed of two main networks: a generator network $G$ and a discriminator network $D$, that are competing against each other. The architecture builds up on the one recently proposed in \cite{yang17}, with conditional input extensions that allow parameters such as force, material and application position to be specified. Broadly, the generator maps the undistorted 3-D model into a deformed 3-D model, conditioned on the supplied parameters. The discriminator is used during training only and is a classifier that determines whether its input is drawn from the ground-truth or the output of the generator. The generator and discriminator are adversarial i.e. they each get better over time. We now describe each network in detail.

\subsection{Generator}
% describe the input...
The network takes as input a voxel grid of size 64$\times$64$\times$64, representing a 3-D point cloud, which is obtained by subsampling the input depth image.

Since a traditional generator from a GAN does not have the ability to map a 3-D point cloud to its latent representation, the generator is implemented as an autoencoder network $E$, that is able to learn a latent representation from input voxel grids.
The encoder enables the network to explore the latent space by
interpolating or making variations on it. By concatenating a \emph{condition} to the latent representation, explicitly controlled
variations can be made as conditional information, such as the size of the force.

In order to facilitate the propagation of local structures in the input voxel grids, the autoencoder has skip connections between the encoder and the decoder.
The encoder has five 3-D convolutional layers with
a bank of 4$\times$4$\times$4 filters with strides of 1$\times$1$\times$1, followed by
a leaky ReLU activation function and a max pooling layer with 2$\times$2$\times$2 filters and 2$\times$2$\times$2 strides. 
%The number of output channels of max pooling layer starts with 64, doubling at each subsequent layer and ends up with 512. 
The encoder is followed by two fully-connected layers which flatten the 3-D representation into a 1-dimensional vector representing the latent encoding. The condition is also encoded as a 1-dimensional vector. The condition encapsulates three properties: the magnitude of the force, the location of the force, and the material. Each of these properties is discretized into a range of values and represented as a one-hot vector.
The condition vector is of the form $(f_2,a_2,m_2)$, where $f_2,a_2,m_2$ are the binary representations of the discretized conditions. In our application scenario, we use 2 bits for the force, 7 for the point of application and 2 for the material.
% in order to map information into latent space. 
The decoder largely follows the inverse of the encoder, composed of five up-convolutional layers which are followed by ReLU activations except for the last layer which is followed by a sigmoid function. 
% \begin{figure}[t]
% \centering
% \includegraphics[width=\columnwidth]{figures/condition.pdf}
% \caption{The condition vector.}
% \label{fig:condition}
% \end{figure}

\subsection{Discriminator}
The discriminator classifies the voxel grids produced by the generator, trying to distinguish whether the predicted outputs are realistic. 
It is composed of five 3-D convolutional layers, each of which has a
bank of 4$\times$4$\times$4 filters with strides of 2$\times$2$\times$2, followed by a
ReLU activation function except for the last layer which is
followed by a sigmoid activation function. %The number of output channels of each layer is the same as that in the encoder part. 
% modify words, now it's the same as bo's paper
The discriminator takes as input pairs of `real' ground truth point clouds and 
`fake' generated clouds, as well as the condition vector.
The condition vector for the discriminator is encoded differently to the generator as 32$\times$32$\times$32 one-hot block masks. This is performed by the Preprocessing kernel, %kernel?
simply replicating the 1-D condition vector on the three axes.
%In particular, the applied force magnitude, point of application %along one axis only???
%and the material type are discretized into 32 values and encoded as a binary block mask.
Including the condition at an early stage of the discriminator makes it possible to model input variations.
We experimentally found that inserting the condition at the second layer gives optimal results.
% explain the discretization

\subsection{Physics engine}
\label{subsec:physicsengine}
%training phase
The physics engine is used to generate ground truth pairs of input point clouds and condition vectors for use in training. 
In this work we used the COMSOL Multiphysics software in order to generate the training voxel grids, but the network is agnostic to the simulator.
The 1-dimensional condition vector is also obtained from the simulator, and it is formed by concatenating the object material and a force vector (point of application and magnitude). Note that a 3-D FEM simulation takes several minutes to several hours depending on the mesh resolution, running on a workstation (CPU - Intel Xeon(R) CPU E5-1603 v3 2.80GHz $\times$ 4). The simulation takes less time if the mesh is coarser, but the risk of not converging becomes higher. 

In our FEM simulations each mesh is composed by around 3000 triangles.

% library of models

%Before the prediction phase, it is possible to estimate per-pixel material composition from RGB images, obtaining a segmented 2.5D image of the body, using available approaches such as the recent Differential Angular Imaging for Material Recognition \cite{dain17}, which was trained on the GTOS (Ground Terrain in Outdoor Scenes) material reflectance database composed of 40 surface classes.

\subsection{Training}
\label{subsec:training}
\begin{figure}[t]
\centering
\includegraphics[width=\columnwidth]{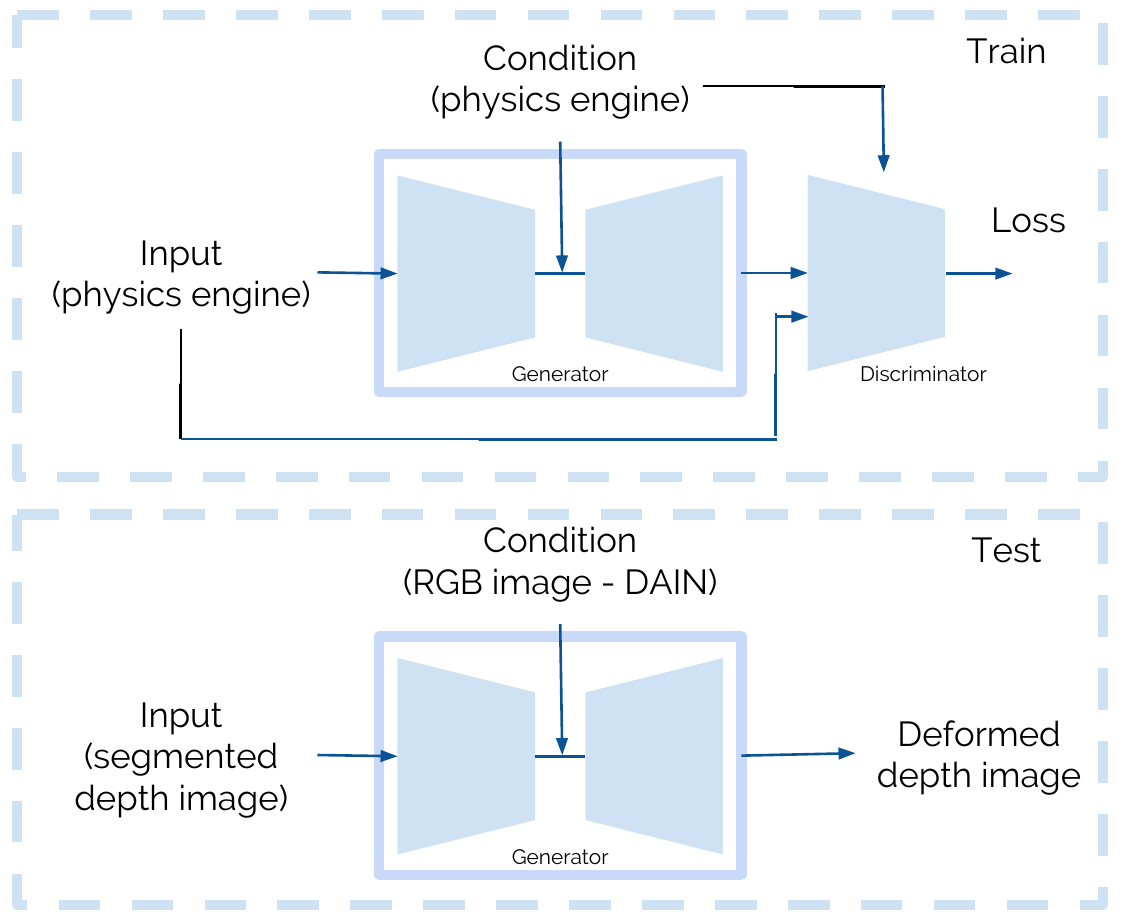}
\caption{Training and testing configurations.}
\label{fig:training}
\end{figure}

Figure \ref{fig:training} shows the training and testing configurations. In the training phase, input and output pairs of voxel grids are generated by the physics simulator.

The adversarial loss $\mathcal{L}^g_{gan}$ is the WP-GAN loss function from \cite{gulrajani17}, with $\lambda=10$.
The reconstruction loss $\mathcal{L}_{AE}$ for the autoencoder $E$ is a specialized form of Binary Cross-Entropy (BCE), as in \cite{brock2016generative}, and is given by:
\begin{equation}
\mathcal{L}_{AE} = -\alpha t \log(o) - (1-\alpha)(1-t) \log(1-o)
\end{equation}
where $t$ is the true occupancy value for each voxel ({0,1}), $o$ is the predicted occupancy value in the range (0,1), $\alpha$ is a weight that balances false positives and false negatives. 

% The adversarial loss for the generator is described by:
% \begin{equation}
% \mathcal{L}^g_{gan} = -E \left[ D(o \vert x) \right]
% \end{equation}
The total loss is:
\begin{equation}
\mathcal{L}^g = \beta \mathcal{L}_{AE} + (1-\beta)\mathcal{L}^g_{gan},
\end{equation}
where $\beta$ is a constant that balances the autoencoder loss and the GAN loss.
% Intuitively, the autoencoder loss is useful for learning the coarse 3D shape of the object, and is important in the first phase of training, while the GAN loss is useful for learning to generate more plausible predictions, in particular the small shape deformations caused by the condition vector. 

The network was trained using the Adam optimizer, with a batch size of 8. 
The learning rate is 0.0005 in the first epoch, and decays to 0.0001.
%training set size...
The network was implemented in Tensorflow and trained on a single Nvidia Titan X GPU. 
% the implementation will be available online along with the training data.

\subsection{Prediction (testing)}
\label{subsec:bodyseg}
In the testing configuration, input is taken from an RGB-D image. To perform a prediction from real-world data, the 2.5D image can be segmented in order to extract the structure to be deformed. This can be performed using existing approaches, such as \cite{qi2016pointnet}. The segmented body can then be upsampled to form a 3-D voxel representation of the object to be traversed. In addition, the material of the body can also be supplied as one of the conditional parameters and obtained from different methods, i.e., the recent Differential Angular Imaging for Material Recognition (DAIN) network \cite{dain17}, trained on the GTOS (Ground Terrain in Outdoor Scenes) material reflectance database composed of 40 surface classes. A prediction on a single Titan X GPU takes under a second, orders of magnitude faster than an FEM simulator.

%%%%%%%%%%%%%%%%%%%%%%%%%%%%%%%%%%%%%%%%%%%%%%%%%%%%%%%%%%%%%%%%%%%%%%%%%%%%%%%%
\section{Experimental evaluation}
\label{sec:experiments}
We test an application of \textsc{Defo-Net} to the problem of predicting deformation of traversable bridges and non-rigid terrain.
In our experiments we use a Turtlebot 2 platform equipped with a Microsoft Kinect camera. The Turtlebot 2 robot weighs 6.3~kg itself and can carry a maximum payload of 5~kg.
The mobile platform has two side wheels and two small castor wheels in the front and back.
In our scenario we assume that the robot is able to segment different materials using DAIN \cite{dain17}. %and segment the point cloud with a deep classifier find cit
For the navigation part we use the algorithms available in the Robot Operating System (ROS).

\subsection{Scenario 1: Safety assessment}
\begin{figure}[h]
\center
\begin{tabular}{cc}
\includegraphics[width=.45\columnwidth,trim={4cm 4cm 4cm 3.5cm},clip]{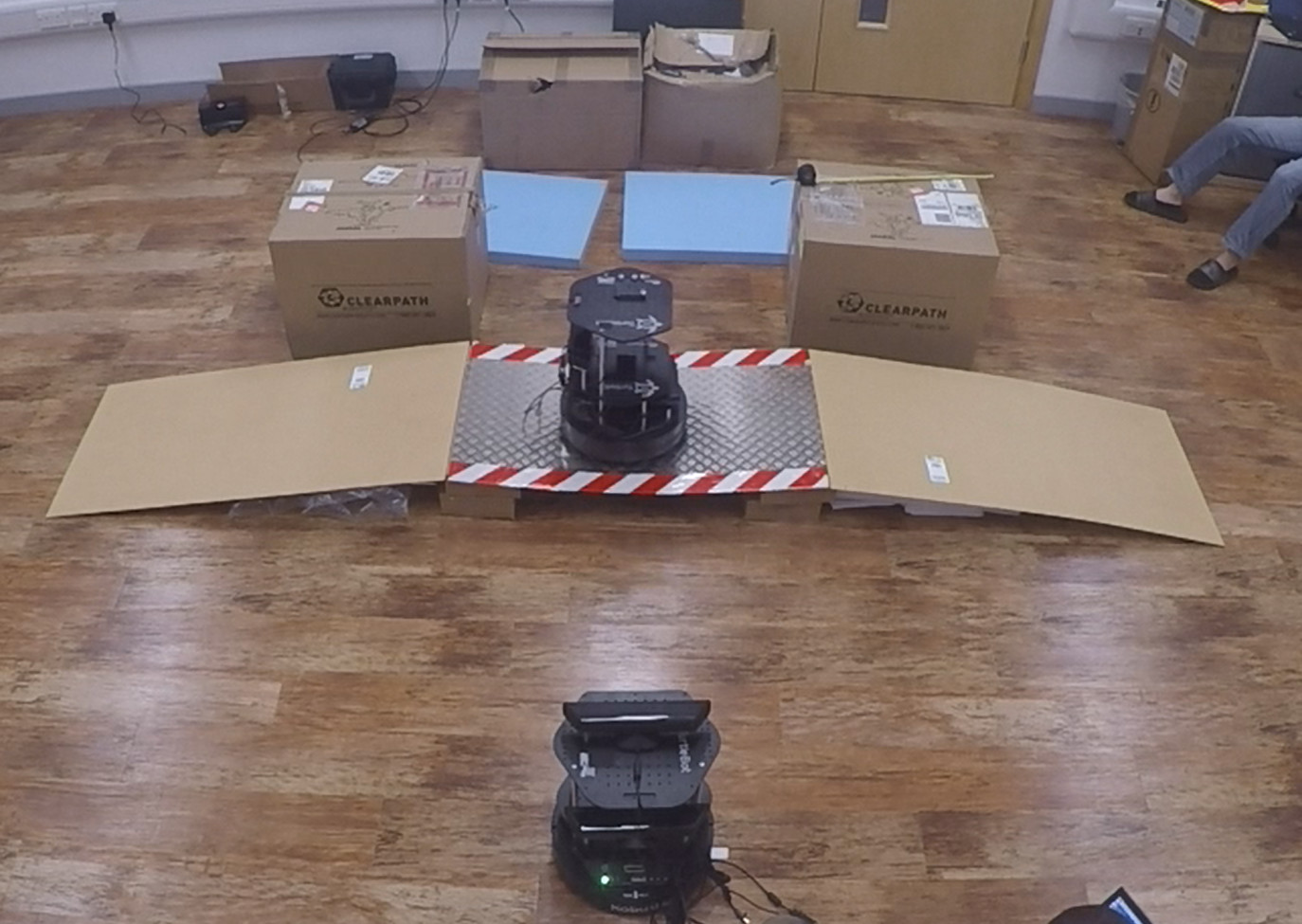} &
\includegraphics[width=.45\columnwidth,trim={4cm 4cm 4cm 3.5cm},clip]{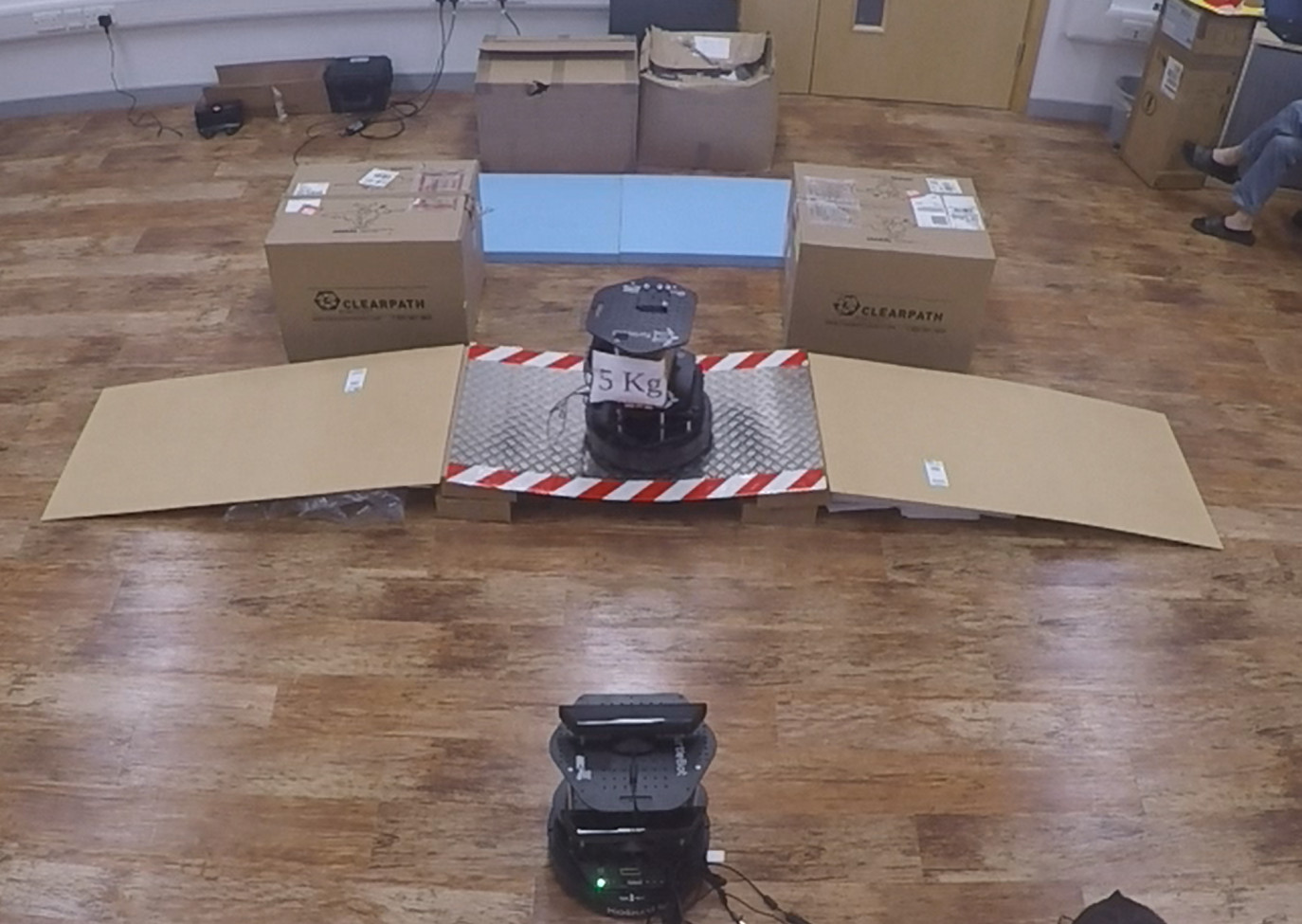} \\
\includegraphics[width=.45\columnwidth,trim={4cm 4cm 4cm 3.5cm},clip]{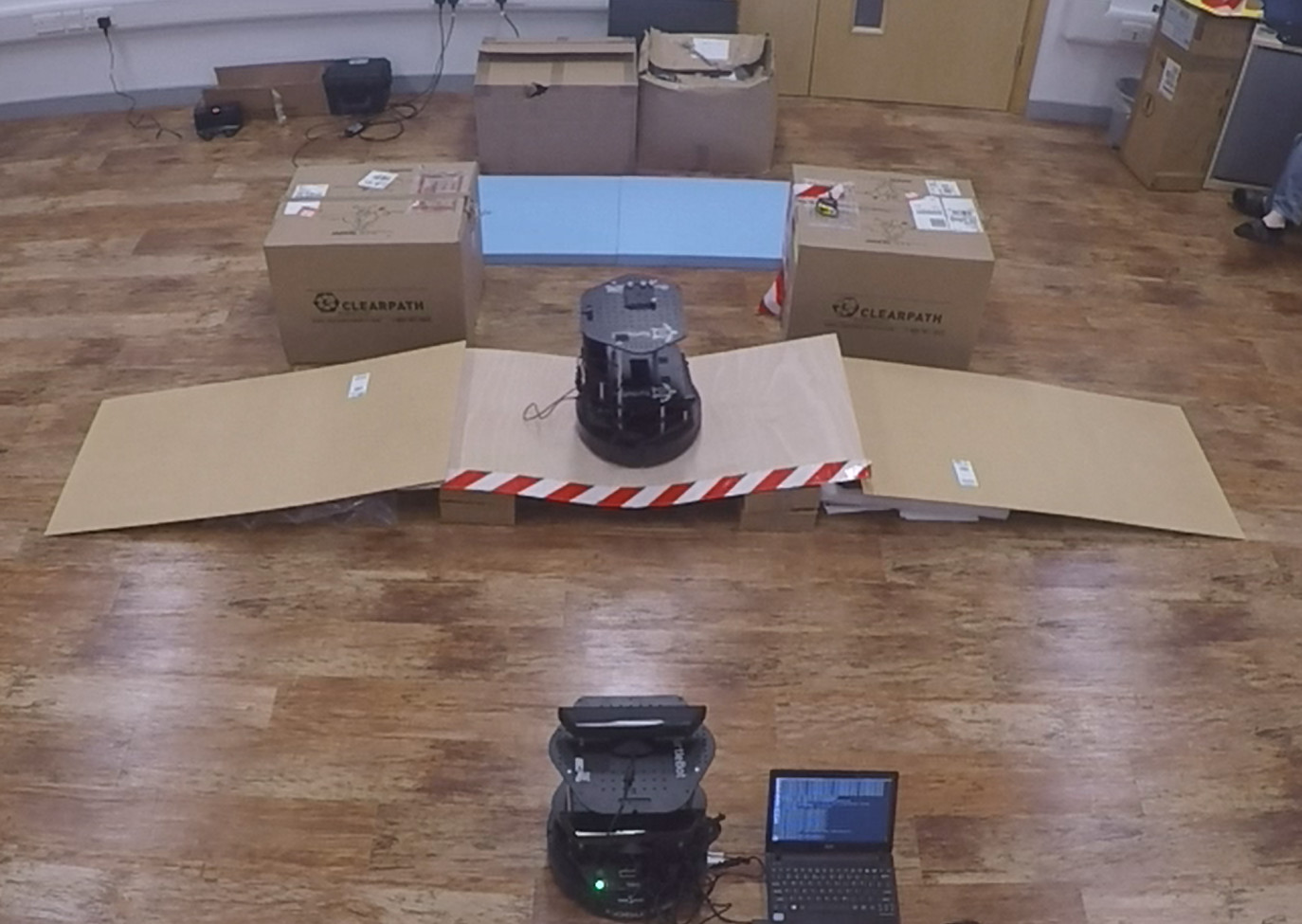} &
\includegraphics[width=.45\columnwidth,trim={4cm 4cm 4cm 3.5cm},clip]{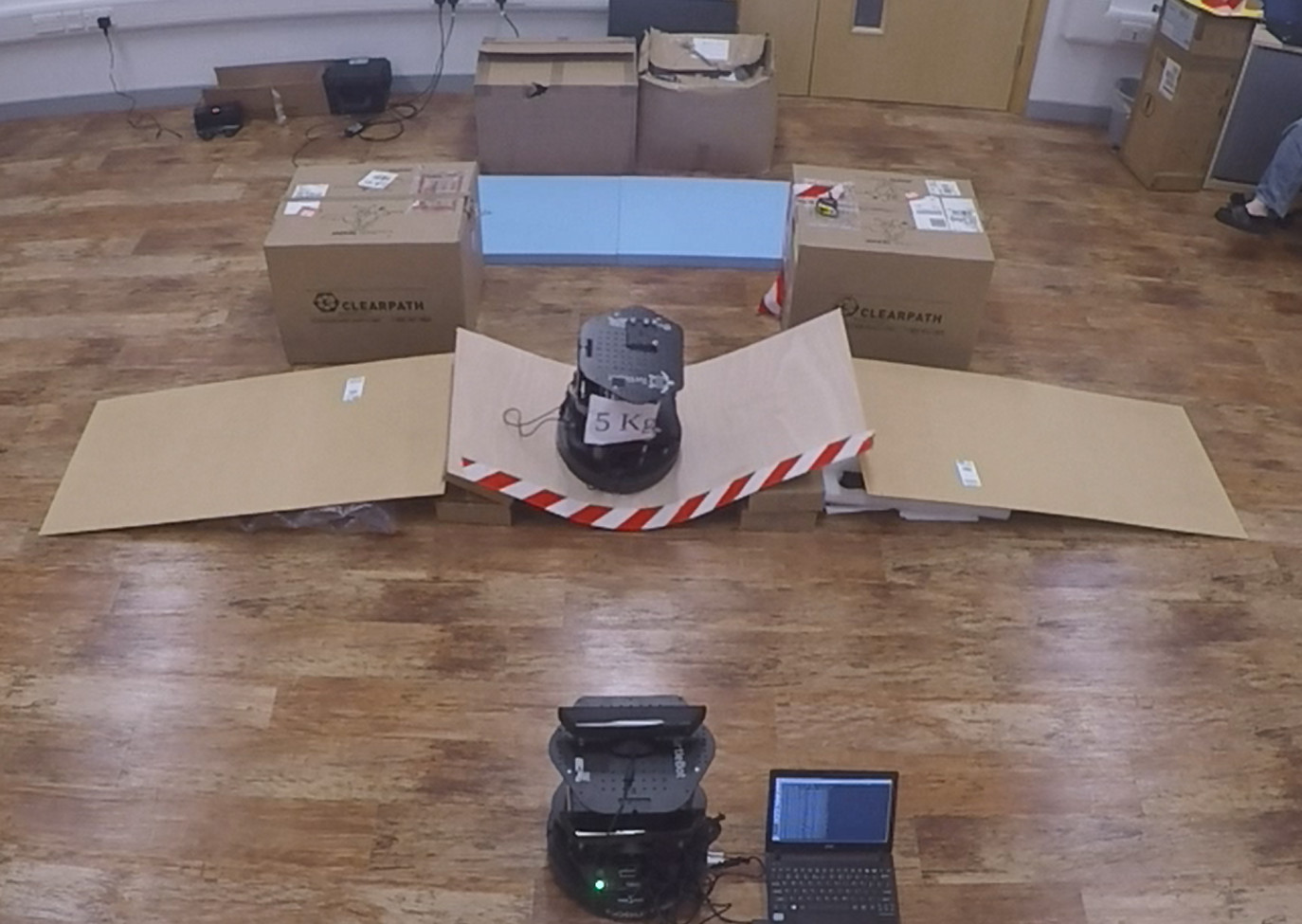}
\end{tabular} 
\caption{The bridge-like scenario.}
\label{fig:bridgescenario}
\end{figure}

In this experiment we consider a scenario in which a robot has to choose whether to cross a bridge and assess whether it is safe to do so by predicting the maximum deformation of the bridge under a known load (payload plus weight of robot).
If the predicted deformation is too large compared with the ground clearance of the robot, the path is considered unsafe.
We show four different cases: the robot without and with a payload crossing a bridge of length 0.6~m made of either plywood or aluminium. In our experiment, as shown in Figure~\ref{fig:bridgescenario}, the thickness of the bridge is similar for the two materials. Note however that it would be possible to add the material thickness as another conditioning variable.

\begin{figure*}[h!]
\centering
\includegraphics[width=\textwidth]{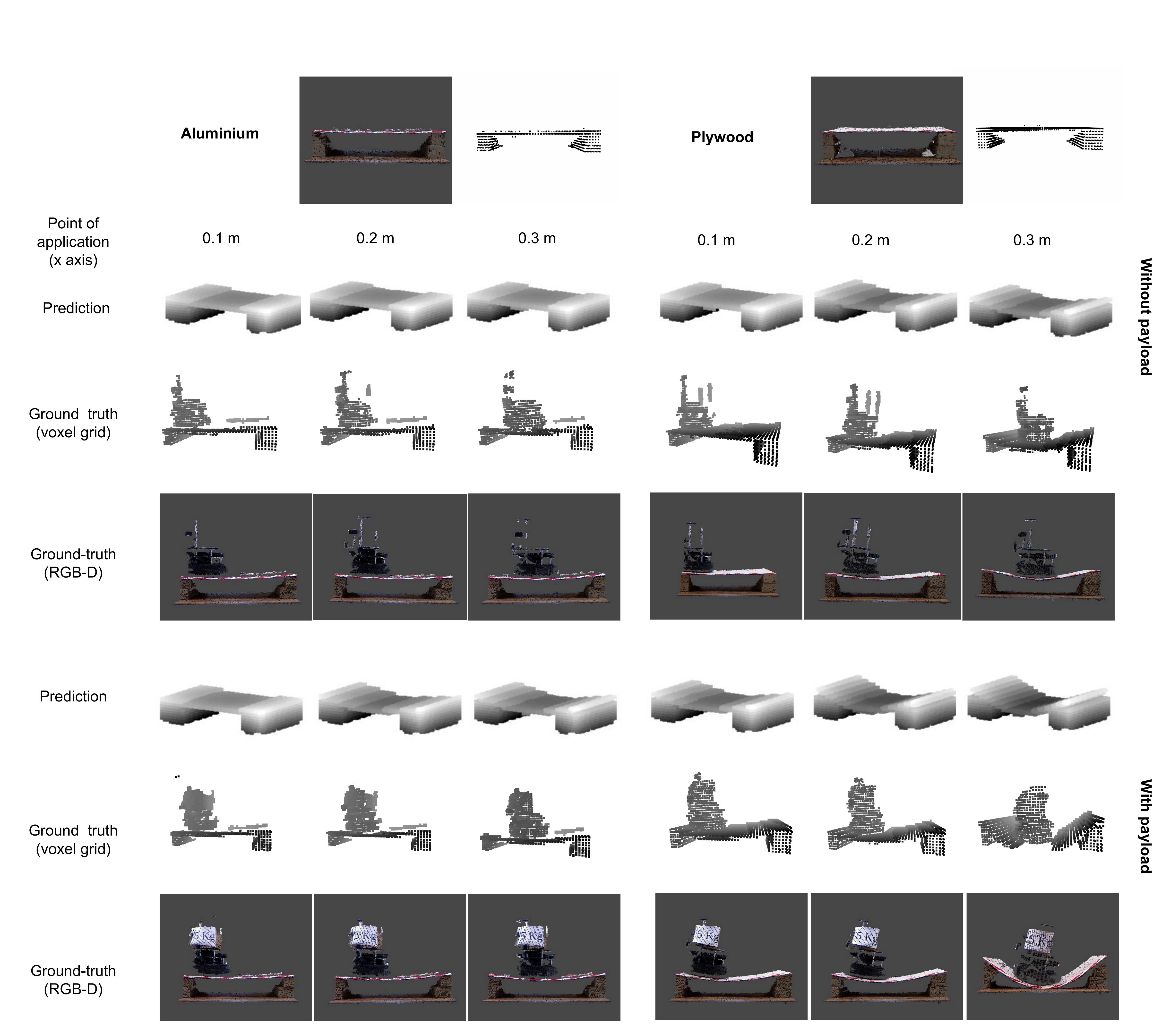}
\caption{Results for the first experiment, where we predict the deformation of a simply supported bridge. We examine two different materials: aluminium (left side of the image) and plywood (right side of the image), with no payload (top half of the image) and with a payload (bottom half of the image). The two input voxel grids for the two materials, along with the 2.5D image, are shown in the top row.}
\label{fig:bridge}
\end{figure*}

The robot first acquires a depth image of the bridge by facing it from the side and extracting it from the depth image.
We compare the predicted deformations with the ground truth from a Kinect camera placed transversally to the bridge, and we let the robot drive over the bridges with and without a payload to obtain the ground-truth.
Figure \ref{fig:bridge} shows the results for the four different cases.
We can see how \textsc{Defo-Net} is able to predict the material deformation under different loads applied to different parts of the bridge. Moreover, it can be seen how the network is able to reconstruct the full 3-D object from the single input depth image.

In this experiment we want to determine if crossing a particular bridge under a particular payload is safe. We define it as safe if the local curvature of the board is below the maximum ground clearance of the robot (0.015 m). From the ground truth in Figure \ref{fig:bridge} we can see that the first three cases are safe, while in the last one (wooden bridge with a payload) the deformation is too large and the robot would become stuck. For comparison, the simulation time for a single deformation is 12m 45s on an Intel Xeon 2.8 GHz CPU while the prediction time is 1~s on a Nvidia Titan X GPU. The path planner can then take this information into account to decide which is the optimal and safest path to a target destination. This experiment demonstrates that by being able to predict the extent of the deformation, a potentially unsafe trajectory can be avoided. However, when unloaded, the robot can safely travel over a wooden bridge - that is to say, the wooden bridge should not always be avoided, only when the robot is fully laden.
In Table \ref{tab:scenario1} we report the  Root-Mean-Square error (RMSE) error for the maximum deformation at different point of the bridge with respect to the ground-truth for each case. In the case of plywood with payload, the real deformation is larger because the bridge collapsed under the weight slipping out of the supports.
%numerical results here

\begin{table}[h]
\centering
\begin{tabular}{llll}
                       & 10cm & 20cm & 30cm \\
\hline \hline                       
Wood - no payload      & 0.1  & 0.4  & 1.5   \\
Aluminium - no payload & 1.2  & 0.7  & 1.3  \\
Wood - payload         & 2.2  & 0.8  & 9.0  \\
Aluminium - payload & 0.8  & 1.4  & 2.7  
\end{tabular}
\caption{RMSE error (cm) between the predicted maximum deformation of the bridge and the ground-truth at different locations.}
\label{tab:scenario1}
\end{table}

\subsection{Scenario 2: Finding the fastest route}
\begin{figure}[t]
\center
\begin{tabular}{c}
\includegraphics[width=.65\columnwidth]{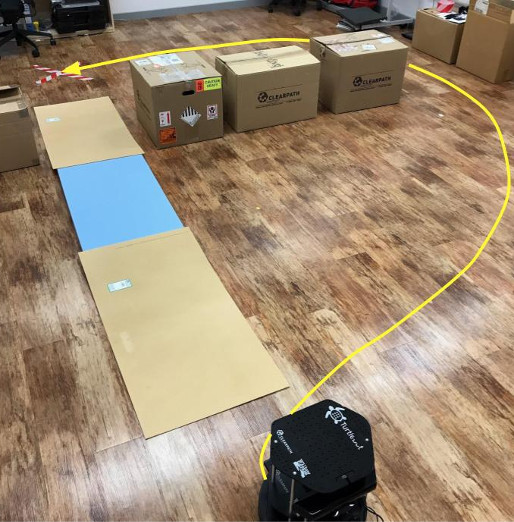}
\end{tabular} 
\caption{The setup for the second scenario. The goal is shown as a red cross in the image. Two paths are available to the goal; the shorter one contains soft ground, the longer one is entirely hard floor. When carrying a payload, the robot avoids travelling over the foam.}
\label{fig:setupwheels}
\end{figure}

We show the performance of the network applied to soft materials like foam, in order to show how \textsc{Defo-Net} is able to learn localized deformations from distributed forces. The experimental setup is shown in Figure \ref{fig:setupwheels}.
In this scenario a robot has to travel from its position to a predefined goal. Two different paths to the goal are available: we let the robot decide between a path containing a soft floor represented in our experiment by a foam board and another path containing only a hard floor but with a longer travel time. The foam board is easily segmented using RGB-D images and can be identified by DAIN as such.

We can predict how the soft floor material is deformed under different payloads for the driven and castor wheels, and based on this prediction make a decision on the maximum speed that the robot can achieve on the soft ground without getting stuck. Based on the maximum speed the robot is able to chose the best path, again based on the ground clearance of the robot, by simply marking the unsafe areas as obstacles for the sake of this experiment. In order to achieve a sufficient resolution, we predict the deformation around the point of contact of each wheel separately.

\begin{figure}[t]
\centering
\includegraphics[width=\columnwidth]{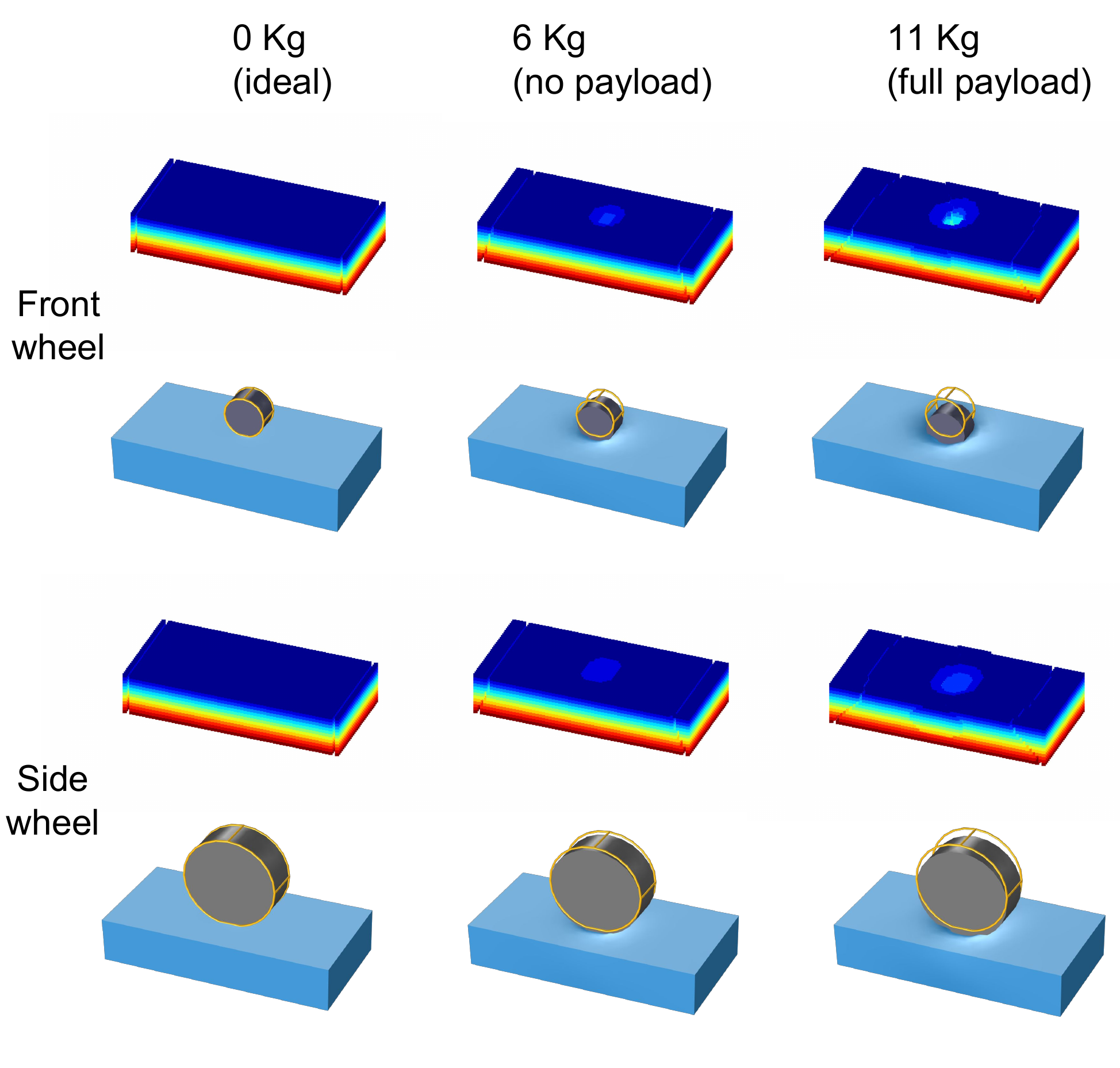}
\caption{Results for the second scenario. First two rows: front castor wheel; last two rows: side wheel. For each payload we show the predicted deformations and the ground truth deformations from the simulator for reference. For visibility, the depth is shown in false colors in the predictions.}
\label{fig:wheels}
\end{figure}

We show the predicted deformations for the front and side wheels in Figure \ref{fig:wheels}, and compare them with the ground truth from the FEM simulation. We can see that the network is able to predict accurate deformations in the presence of different contact areas.  More, importantly, it is able to correctly predict an excessive deformation of the foam for the castor wheels with the full payload. In this case, the robot would be unable to steer and could get stuck. Using this predicted information, the robot can safely avoid the foam when carrying a full payload. Note however, that when unloaded, the robot can choose the shortest path, i.e. over the foam. Thus, it can be seen how knowledge of the deformation of the terrain can greatly assist in path planning.

\subsection{Scenario 3: Generalisation ability}
\begin{figure}[h!]
\centering
\includegraphics[width=\columnwidth]{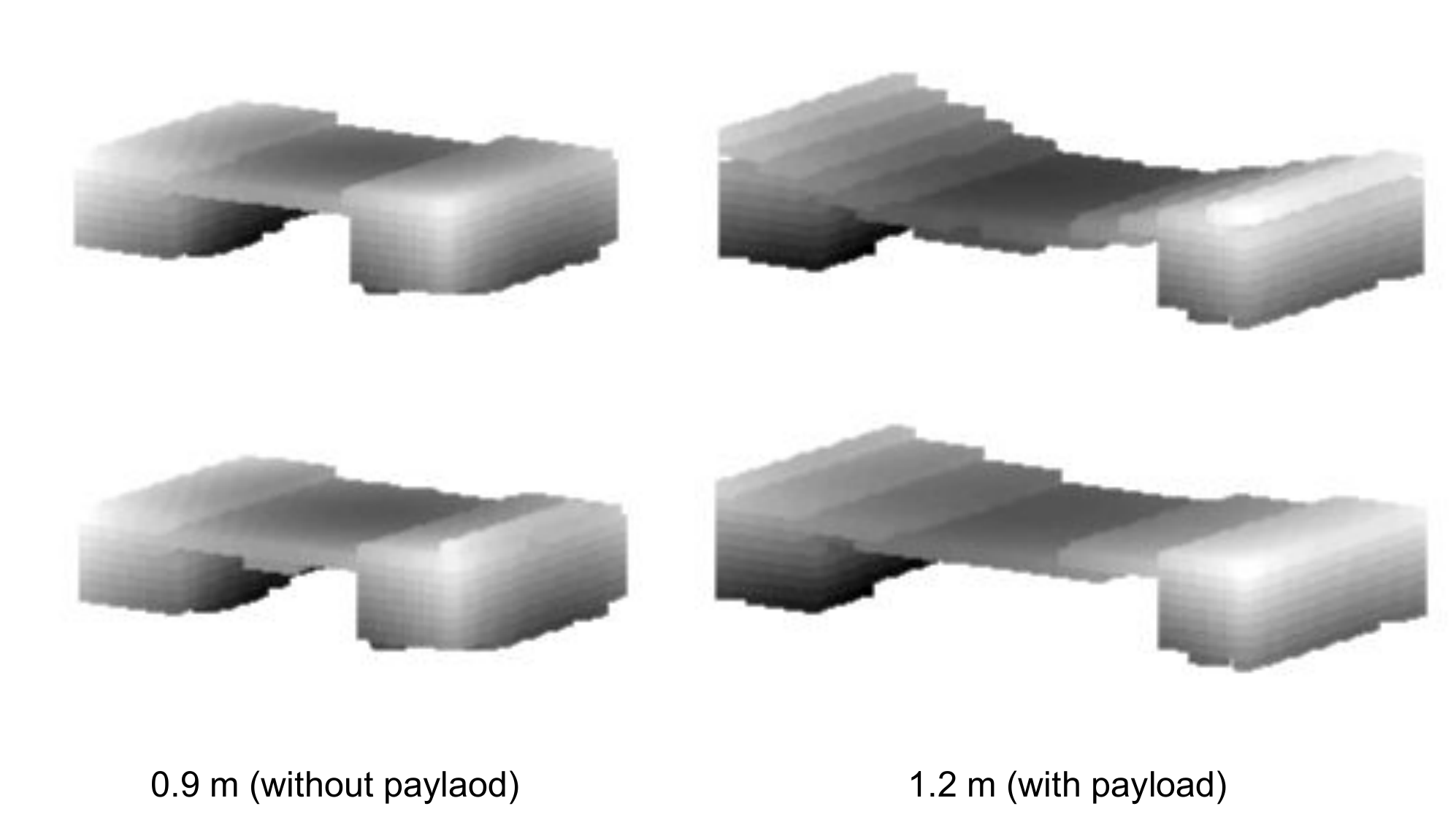}
\caption{Results for the third scenario. The top row shows the predicted 3-D shapes, while the bottom row shows the ground-truth.}
\label{fig:generalisation}
\end{figure}

Finally, in order to test the generalisation abilities of our network, we further train the network on the same bridge-like structure on seven different lengths ranging from 0.8m to 1.3m.
Then we test the predicted deformation on two lengths that were chosen at random and are not part of the training set (namely 0.9m and 1.2m), with two different forces acting on the middle of the bridge (robot without payload and with payload).
Figure \ref{fig:generalisation} shows the predicted deformations along with the ground-truth from the FEM simulator for reference. The results show how the generative network is able to reconstruct unseen voxel grids from partial depth images of the test set.
The absolute error on the point of maximum deformation of the predictions with respect to the ground-truth is one voxel (2.2cm) for the bridge of length 0.9m and 2 voxels (4.4cm) for the 1.2m bridge.   

This experiment demonstrates that \textsc{Defo-Net} is able to generalize to unseen, yet related, scenarios. This is of key importance as it is impossible to show the network every possible structure that could be encountered in the real-world. However, the network has learned the relationship between structure, material and force to predict deformations.

\section{Conclusions}
\label{sec:conclusions}
We presented \textsc{Defo-Net}, a generative network for predicting 3-D deformations of bodies extracted from single RGB-D images using invertible conditional GANs.
We applied the network to the problem of safe and optimal navigation of robots carrying payloads over different obstacles and ground floor materials.
Experimental results in a real robotic scenario showed the generalisation potential of the approach to previously unseen body configurations. %Future work will be devoted to investigating diverse applications such as deformation prediction for non-rigid robotic arms, compliant manipulation of soft objects, and incorporating the approach in a path planning framework, as well as unsupervised learning.
More importantly, the prediction can be up to 2-3 orders of magnitude faster than an FEM simulation, making it suitable for real-time navigation. Although this work has set out a new approach towards tackling an active research area, a number of extensions could be considered. 

The first is to build models of more realistic and complex structures. Moreover, we have only considered a small subset of materials (e.g. wood, foam, aluminium), but it would be interesting to see how to treat say plywood differently from solid wood. It would also be interesting to see if the proposed approach could accurately model non-homogeneous media such as pebbles and sand and approximate deformations due to grain interactions.

The second is to consider non-rigid (e.g. soft-body) robots interacting with non-rigid objects e.g. a soft robot folding a blanket or closing curtains.

Another direction would be to investigate dynamic deformations such as those that would be caused by a robot applying a time-varying force to a non-rigid object, rather than the static loads considered here.

A further area of research would be to consider alternative world representations. Although voxels map neatly to RGB-D images, they are not a natural representation of deformable solids. Non-Euclidean approaches based on graph geometry and manifolds would be a better fit to the polygonal meshes used in FEM simulation. 

\section{Acknowledgements}
This work has been partly supported by the EPSRC Program Grant ``Mobile Robotics: Enabling a Pervasive Technology of the Future (GoW EP/M019918/1)". 
%The authors would like to thank CSC for the scholarship supporting Zhihua Wang's oversea study. 

\bibliographystyle{IEEEtran}
\bibliography{IEEEabrv,biblio}

\end{document}